\documentclass[journal]{IEEEtran}

\usepackage{graphicx}
\usepackage{amssymb,amsmath,epsfig}
\usepackage{algorithm}
\usepackage{algorithmic}
\usepackage{booktabs}
\usepackage{mathrsfs}
\usepackage{color}
\usepackage{textcomp}
\usepackage{bbding}
\usepackage{pifont}
\usepackage{wasysym}
\usepackage{amssymb}
\usepackage{subcaption}

\begin{document}
%
\title{MSCFNet: A Lightweight Network With Multi-Scale Context Fusion for Real-Time Semantic Segmentation}
%
%
%
\author{Guangwei Gao,~\IEEEmembership{Member,~IEEE,}
        Guoan Xu,
        Yi Yu,~\IEEEmembership{Member,~IEEE,}
        Jin Xie,~\IEEEmembership{Member,~IEEE,}
        Jian Yang,~\IEEEmembership{Member,~IEEE,}
        and Dong Yue,~\IEEEmembership{Fellow,~IEEE}
\thanks{Manuscript received XXXX; revised XXXX; accepted XXXX. This work was supported in part by the National Key Research and Development Program of China under Project nos. 2018AAA0100102 and 2018AAA0100100, the National Natural Science Foundation of China under Grant nos. 61972212, 61772568 and 61833011, the Natural Science Foundation of Jiangsu Province under Grant no. BK20190089, the Six Talent Peaks Project in Jiangsu Province under Grant no. RJFW-011, and Open Fund Project of Provincial Key Laboratory for Computer Information Processing Technology (Soochow University) (No. KJS1840).~\textit{(Corresponding author: Guangwei Gao and Guoan Xu.)}}
\thanks{G. Gao is with the Institute of Advanced Technology, Nanjing University of Posts and Telecommunications, Nanjing 210023, China, with the Digital Content and Media Sciences Research Division, National Institute of Informatics, Tokyo 101-8430, Japan, and also with the Provincial Key Laboratory for Computer Information Processing Technology, Soochow University, Suzhou 215006, China (e-mail: csggao@gmail.com).}
\thanks{G. Xu is with the College of Automation \& College of Artificial Intelligence, Nanjing University of Posts and Telecommunications, Nanjing 210023, China (e-mail: xga\_njupt@163.com).}
\thanks{Y. Yu is with the Digital Content and Media Sciences Research Division, National Institute of Informatics, Tokyo 101-8430, Japan (e-mail: yiyu@nii.ac.jp).}
\thanks{J. Xie and J. Yang are with the School of Computer Science and Technology, Nanjing University of Science and Technology, Nanjing 210094, China (e-mail: csjxie@njust.edu.cn; csjyang@njust.edu.cn).}
\thanks{D. Yue is with the Institute of Advanced Technology, Nanjing University of Posts and Telecommunications, Nanjing 210023, China, and also with the College of Automation \& College of Artificial Intelligence, Nanjing University of Posts and Telecommunications, Nanjing 210023, China (e-mail: medongy@vip.163.com).}
}

\markboth{IEEE Transactions on Intelligent Transportation Systems}%
{Shell \MakeLowercase{\textit{et al.}}: Bare Demo of IEEEtran.cls for IEEE Journals}
%

\maketitle

\begin{abstract}
In recent years, how to strike a good trade-off between accuracy, inference speed, and model size has become the core issue for real-time semantic segmentation applications, which plays a vital role in real-world scenarios such as autonomous driving systems and drones. In this study, we devise a novel lightweight network using a multi-scale context fusion (MSCFNet) scheme, which explores an asymmetric encoder-decoder architecture to alleviate these problems. More specifically, the encoder adopts some developed efficient asymmetric residual (EAR) modules, which are composed of factorization depth-wise convolution and dilation convolution. Meanwhile, instead of complicated computation, simple deconvolution is applied in the decoder to further reduce the amount of parameters while still maintaining the high segmentation accuracy. Also, MSCFNet has branches with efficient attention modules from different stages of the network to well capture multi-scale contextual information. Then we combine them before the final classification to enhance the expression of the features and improve the segmentation efficiency. Comprehensive experiments on challenging datasets have demonstrated that the proposed MSCFNet, which contains only 1.15M parameters, achieves 71.9\% Mean IoU on the Cityscapes testing dataset and can run at over 50 FPS on a single Titan XP GPU configuration.
\end{abstract}


%
\IEEEpeerreviewmaketitle

\section{Introduction}
\label{sec1}

\IEEEPARstart{A}{utonomous} driving technology has been widely studied to enhance the driving experience and relieve traffic pressure~\cite{lu2019cognitive,gao2020cross}. With the help of cameras, it becomes easier to perceive and understand the surrounding environment~\cite{zhao2017multisensor,gao2017learning,gao2020constructing}. Semantic segmentation, which aims at assigning a category to each pixel for the given image, is a challenging research topic in the field of computer vision. Recently, deep convolutional neural networks (DCNNs)~\cite{he2016deep,rawat2017deep,gao2020hierarchical} have shown their impressive capabilities on image classification with high resolution. Especially the fully convolutional network (FCN)~\cite{shelhamer2017fully}, which is a pioneer CNN for semantic segmentation task. The encoder-decoder network has also become a popular structure for solving the segmentation problem. Although achieving remarkable results, most of the previous networks~\cite{lu2017wound,chen2017deeplab,yang2018denseaspp,fu2019dual,sun2019high} ignored the segmentation efficiency, namely, their calculation and storage requirements are so high that it is difficult to meet the demands of real-world applications where information needs to interact quickly with the environment. Meanwhile, electric equipments such as robotics, cellphones, and telemedicine, etc, having small memory capacity and limited computational cost, cannot support the enormous complex algorithms.	

Therefore, it is a primary trend to design lightweight and efficient networks to overcome the above problems. The smaller-scale network means a faster inference speed and less redundancy~\cite{lan2020madnet}. However, most of the existing real-time research works~\cite{paszke2016enet,badrinarayanan2017segnet,romera2017erfnet,mehta2019espnetv2} mainly focus on shallowing the networks and reducing parameters to shorten the time-consuming at the expense of model accuracy.

\begin{figure*}[t]
	\centering
	\includegraphics[width=17.7cm]{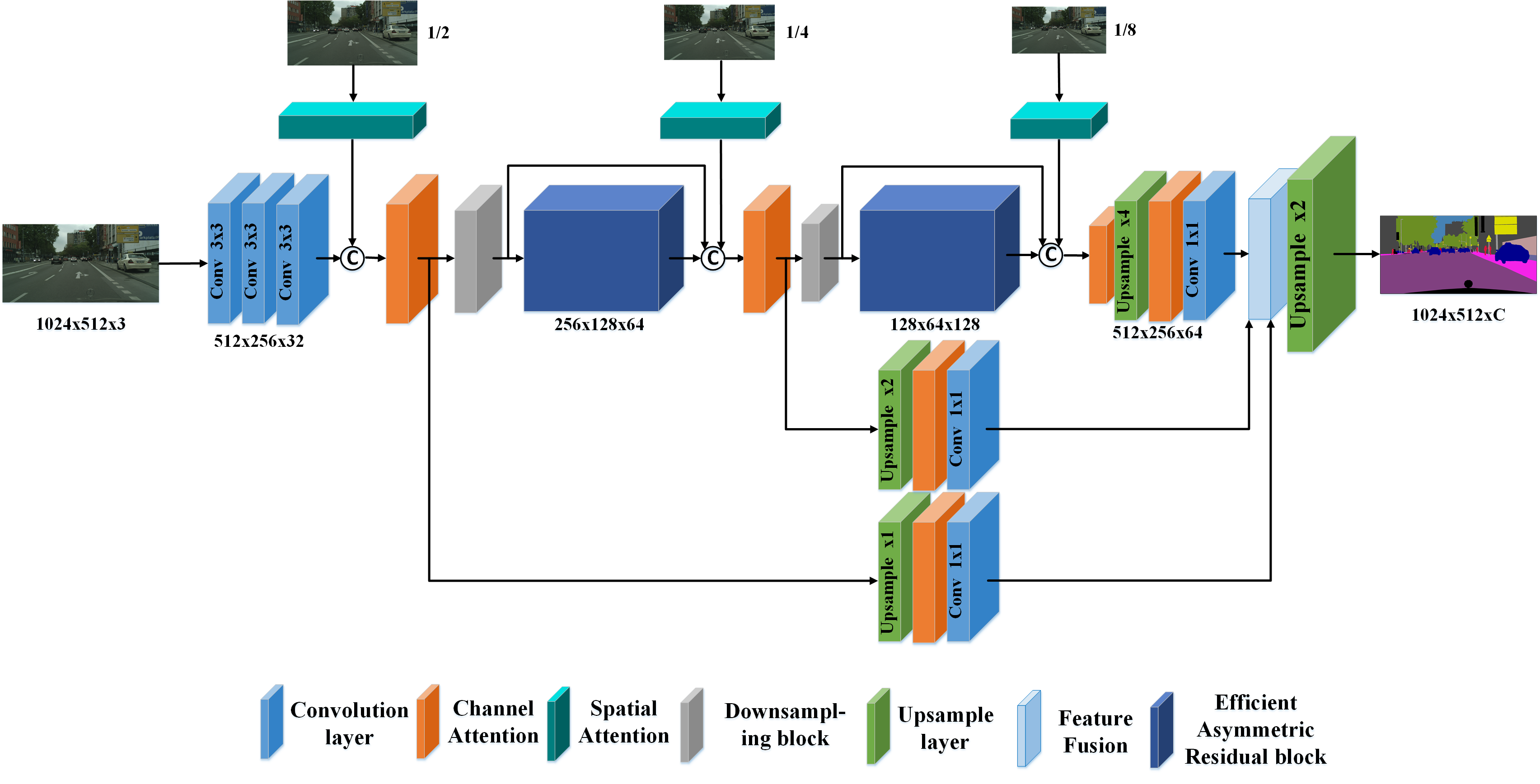}
	\caption{The procedure of our proposed MSCFNet. The sizes (width, height, and channel) of the intermediate features are given in the process of network. ``C'' denotes the concatenation, ``$ \times 1$, $ \times 2$ and $ \times 4$'' mean the upsampling factor, ``$\rm{1/2}$, $\rm{1/4}$ and $\rm{1/8}$'' indicate the ratio of the original image scale. (Best viewed in color)}
	\label{Figure 1}
\end{figure*}

In this work, we devise a novel lightweight and efficient network, called multi-scale context fusion network (MSCFNet), to get a better balance between the accuracy and efficiency for real-time semantic segmentation task. Like most of the previous works, our proposed model also explores an asymmetric encoder-decoder structure. As presented in Fig.~\ref{Figure 1}, our MSCFNet has multiple branches with efficient attention mechanisms from different stages of the network, containing multi-scale contextual information for segmentation purpose. Although a small number of parameters and calculations have been added, the general execution improves a lot. The core unit of our MSCFNet is an efficient asymmetric residual (EAR) module with dilated factorized depth-wise separable convolution, which allows us to extract attentive and cooperative feature information on a large receptive field efficiently and quickly.

Our main contributions can be listed as three-fold:
\begin{itemize}
\item We devise an efficient asymmetric residual (EAR) module and construct a lightweight semantic segmentation network with a multi-scale context fusion scheme, which fuses the attentive features adaptively, contributing to the efficiency and effectiveness of the segmentation task.
\item Short-range and long-range connections with efficient spatial and channel attention presented in our method facilitate the local and contextual information interaction greatly, contributing to the improvement of the performance.
\item Our network achieves prominent performance on both Cityscapes and CamVid datasets without any other data augment skills. It has 1.15M model size, while achieves a mean intersection over union (mIoU) of 71.9\% and 69.3\% on Cityscapes and CamVid datasets, respectively.
\end{itemize}

\section{Related Work}
\label{sec2}

\subsection{Factorization convolution}
\label{sec21}

Factorization convolution is often used to improve the efficiency where a traditional two-dimensional convolution is replaced by two one-dimensional convolutions. Xception~\cite{chollet2017xception} and MobileNet~\cite{howard2017mobilenets} applied depth-wise separable convolution, where each input channel and each filter kernel is divided into a group, which operates individually. ERFNet~\cite{romera2017erfnet}, DABNet~\cite{li2019dabnet}, and LEDNet~\cite{wang2019lednet} decomposed a $3 \times 3$ convolution into a $3 \times 1$ and a $1 \times 3$ convolution. They are all beneficial from the factorization convolution, which can reduce the amount of computational burdens.

\subsection{Dilation convolution}
\label{sec22}

Dilation convolution is used to insert zeros between two adjacent kernel values of the standard convolution to achieve the purpose of enlarging the receptive field without adding the parameters. For example, DeepLab series~\cite{chen2017deeplab,chen2018encoder,chen2017rethinking} suggested a spatial pyramid pooling module that adopts various dilation rates arranged as a pyramid. LEDNet~\cite{wang2019lednet} designed a split-shuffle-non-bottleneck (SS-nbt) module using the dilation convolution to construct an asymmetric encoder-decoder architecture. Dilation8~\cite{yu2015multi} proposed a multi-scale context aggregation network by dilated convolutions. EDANet~\cite{lo2019efficient} incorporated dilated convolution and dense connection to attain high efficiency.

\subsection{Lightweight segmentation networks}
\label{sec23}

Lightweight segmentation networks are eagerly required to attain the desired balance between the prediction accuracy and the related inference efficiency~\cite{zhang2020farsee,yang2020small,yang2020ndnet,miclea2019real,han2020using,8392426,zhou2019high,li2019dfanet}. ENet~\cite{paszke2016enet} was the first lightweight architecture used in real-time applications, which trimmed the amount of the convolution filters to decrease the calculation. ESPNet~\cite{mehta2018espnet} proposed an effective spatial pyramid module, which can collect multi-scale contextual information. ICNet~\cite{zhao2018icnet} used a strategy called image cascade to improve the segmentation efficiency. BiseNet~\cite{yu2018bisenet} introduced two branches, one is to retain shallow spatial information and the other is to extract deep contextual information. LEDNet~\cite{wang2019lednet} showed the benefit brought by the channel split and channel shuffle operations. Although these networks have made a relatively satisfactory trade-off between performance and speed, there is still adequate room for further promotion.

\begin{figure}[t]
	\centerline{\includegraphics[width=9cm]{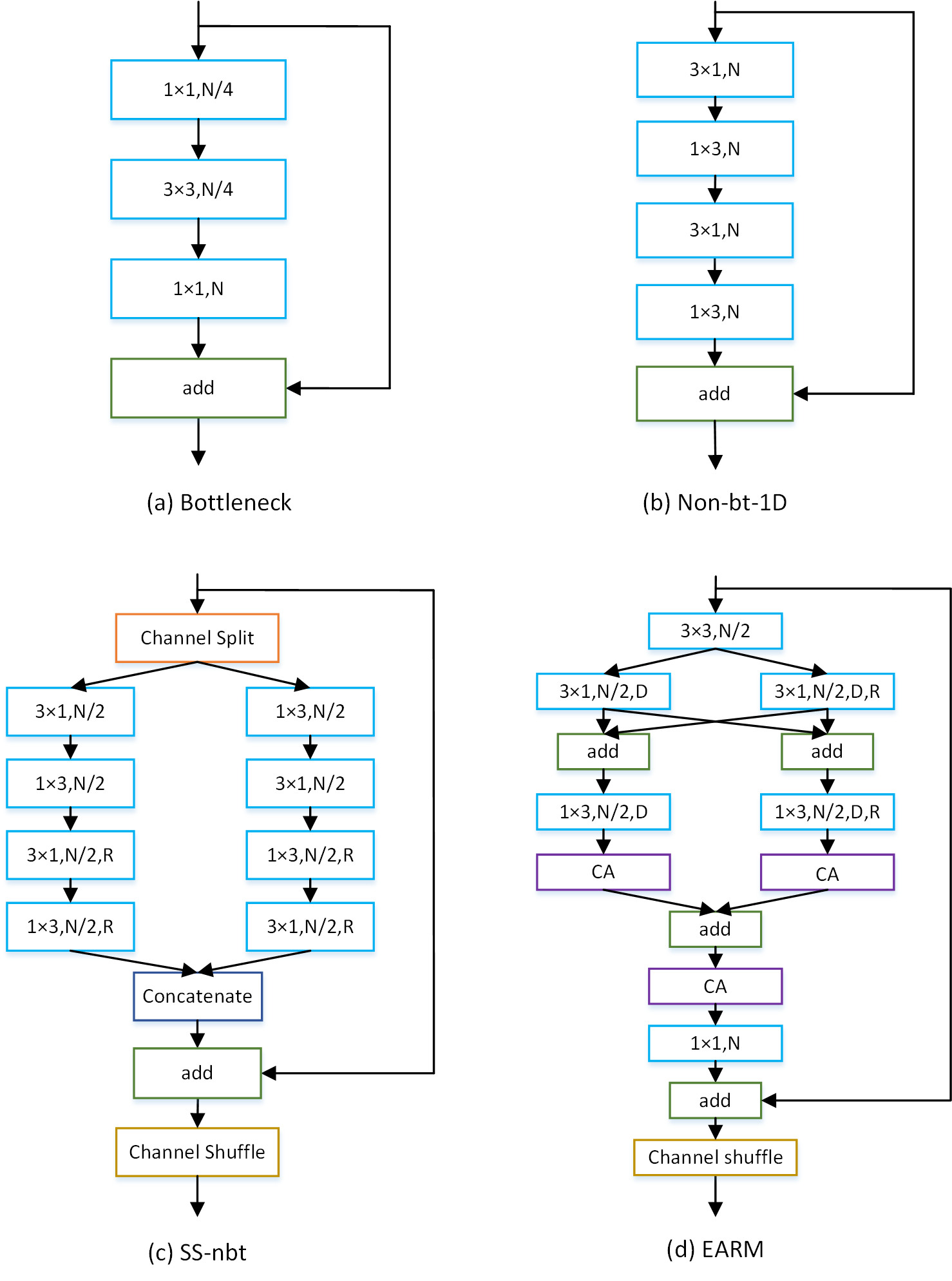}}
	\caption{Comparisons of various types of residual modules. ``N'' is the number of the output channels, ``R'' represents the dilation rate of kernel and ``D'' indicates a depth-wise convolution. (a) Bottleneck~\cite{paszke2016enet}. (b) Non-bt-1D of ERFNet~\cite{romera2017erfnet}. (c) SS-nbt of LEDNet~\cite{wang2019lednet}. (d) Our EAR module.}
	\label{Figure 2}
\end{figure}

\subsection{Attention mechanism}
\label{sec24}

Attention mechanism has been broadly adopted in the field of pattern recognition and computer vision. Its essence is to imitate the human visual mechanism to learn a weight distribution of the image features and apply these weights to the original features. CCNet~\cite{huang2019ccnet} devised an efficient criss-cross attention module to capture the image dependencies. GCNet~\cite{cao2019gcnet} and ANN~\cite{zhu2019asymmetric} further observed the non-local attention mechanism and achieved promising performance for the semantic segmentation task. DANet~\cite{fu2019dual} used the channel and spatial attention tricks simultaneously to model the semantic inter-dependencies. Some of the above works performed sophisticated matrix multiplication on the pixel level, which is not suitable for lightweight applications.

SENet~\cite{hu2018squeeze}, which is a lightweight threshold mechanism, has been widely used to model the correlation of all channels. It first employs a global average pooling to squeeze the global spatial information into channel descriptors and then uses two fully connected layers to capture cross-channel interaction. GENet~\cite{hu2018gather} introduced a pair of operators, consisting of gathering feature responses from a large scale and exciting this information to local features. CBAM~\cite{woo2018cbam} sequentially inferred attention maps along spatial dimension and channel dimension separately, and then the input feature maps are multiplied to the attention maps for adaptive feature refinement. GSoP-Net~\cite{gao2019global} introduced higher-order representation across from lower to higher layers to effectively explore those statistical information. By dissecting the mechanism in SENet, ECANet~\cite{wang2020eca} proposed an effective yet efficient cross-channel interaction scheme avoiding channel dimensionality reduction. It performs well on the tasks of object detection and image classification in terms of parameters and computations.

In contrast to the above approaches, in our method, we design a lightweight semantic segmentation network with a contextual fusion structure to speed up the network training, reduce the model size, and meanwhile ensure the effectiveness of the final inference results. Regarding the popular attention mechanism, we inject spatial and channel attention based on the different conditions of the network at different stages. Simultaneous consideration of these schemes boosts the final segmentation accuracy.

\begin{figure}[t]
	\centerline{\includegraphics[width=7cm]{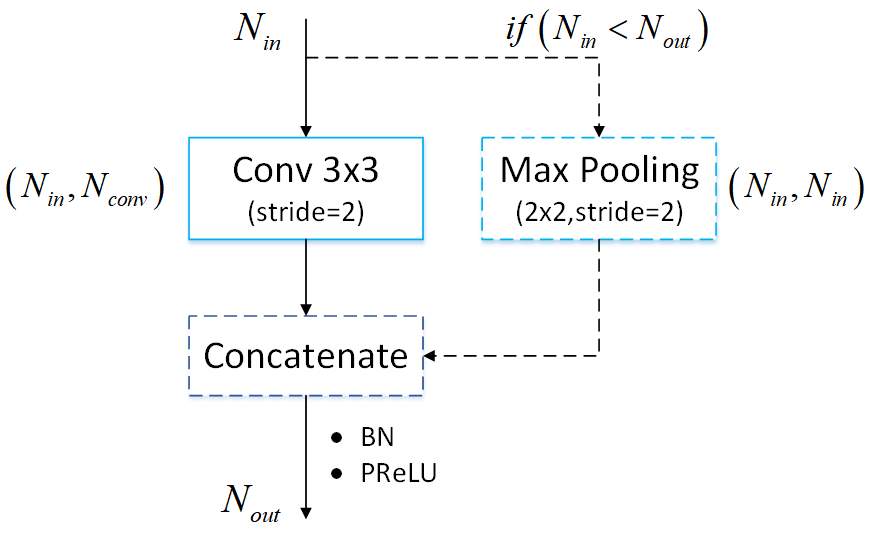}}
	\caption{Downsampling block structure. $N_{in}$: input channel, $N_{out}$: output channel, $N_{conv}$: output channel after convolution, $BN$: Batch Normalization.}
	\label{Figure 3}
\end{figure}

\section{Methodology}
\label{sec3}

In this section, we first illustrate our efficient asymmetric residual (EAR) module and then introduce the efficient attention and context fusion modules. Finally, we elaborate the whole network architecture which consists of an initial block, three input injection modules, two downsampling blocks, two EAR blocks, and two context branches. The entire structure of the proposed MSCFNet is given in Fig.~\ref{Figure 1}.

\subsection{EAR Module}
\label{sec31}

Lightweight networks have witnessed a lot of residual designs (See Fig.~\ref{Figure 2}). Inspired by these designs, we devise the efficient asymmetric residual (EAR) module with their common advantages to achieve a better result under the circumstance of limited computational capacity. Our EAR module is shown in Fig.~\ref{Figure 2} (d). Firstly, the number of input channels is reduced to half by a $3 \times 3$ convolution at the bottleneck. The reason why we use a $3 \times 3$ convolution instead of a $1 \times 1$ convolution which has fewer parameters is that when using $1 \times 1$ convolution, the residual block must construct deeper for a larger receptive field capturing more contextual information, the computational cost and memory requirements must increase. The following is a two-branch structure. One branch applies factorization convolution to depth-wise convolution so that it can collect local and short-range feature information. Specifically, a standard $3 \times 3$ depth-wise convolution is divided into a $3 \times 1$ convolution and a $1 \times 3$ convolution. They would have the same size of the receptive field, while the latter has a fewer number of parameters. Another branch adopts dilation convolution enlarging receptive field to the factorization depth-wise convolution to capture complex and long-range feature information. To avoid the gridding artifacts, we use different dilation rates in different EAR modules, which are not integer powers of 2. 

For the sake of sharing information for different branches, we put the feature interaction operations between $3 \times 1$ and a $1 \times 3$ convolution in the two branches. In such a way, the contextual information extracted by the two branches can complement each other. The feature maps from each branch are then sent to the channel attention module for better extracting discriminative features. And then, the two low-dimension branches are fused and fed into the channel attention module for the same purpose. Following is a $1 \times 1$ point-wise convolution to recover the related channels of the feature maps. Finally, For the sake of evading the drawback of information independence between channels caused by depth-wise convolution, we explore a channel shuffle followed the combination of the output of $1 \times 1$ point-wise convolution and the input to facilitate the channel information exchanging and sharing. The above operations can be expressed as follow:

\begin{equation}
{x_b} = {C_{3 \times 3}}\left( {\rho \left( {{x_{EARin}}} \right)} \right),
\label{eq1}
\end{equation}
\begin{equation}
{y_1} = CA\left( {{C_{1 \times 3}}\left( {{C_{3 \times 1}}\left( {{x_b}} \right) + {C_{3 \times 1,d}}\left( {{x_b}} \right)} \right)} \right),
\label{eq2}
\end{equation}
\begin{equation}
{y_2} = CA\left( {{C_{1 \times 3,d}}\left( {{C_{3 \times 1,d}}\left( {{x_b}} \right) + {C_{3 \times 1}}\left( {{x_b}} \right)} \right)} \right),
\label{eq3}
\end{equation}
\begin{equation}
{y_{EARout}} = S\left( {{C_{1 \times 1}}\left( {CA\left( {\rho \left( {{y_1} + {y_2}} \right)} \right)} \right) + {x_{EARin}}} \right),
\label{eq4}
\end{equation}
where $x_{EARin}$ and $y_{EARout}$ represent the input and output of the EAR module, $x_b$ is the output of $3 \times 3$ convolution, $y_1$ and $y_2$ are the outputs of two branches in EAR module, $C_{m \times n}$ denotes convolution operation with the kernel size $m \times n$, $d$ is the dilation rate, $\rho$ is the PReLU nonlinear activation function, $S$ means channel shuffle operation.

\begin{figure}[t]
    \centering
        \includegraphics[width=10 cm, height=6.8 cm, trim=34 16 0 25]{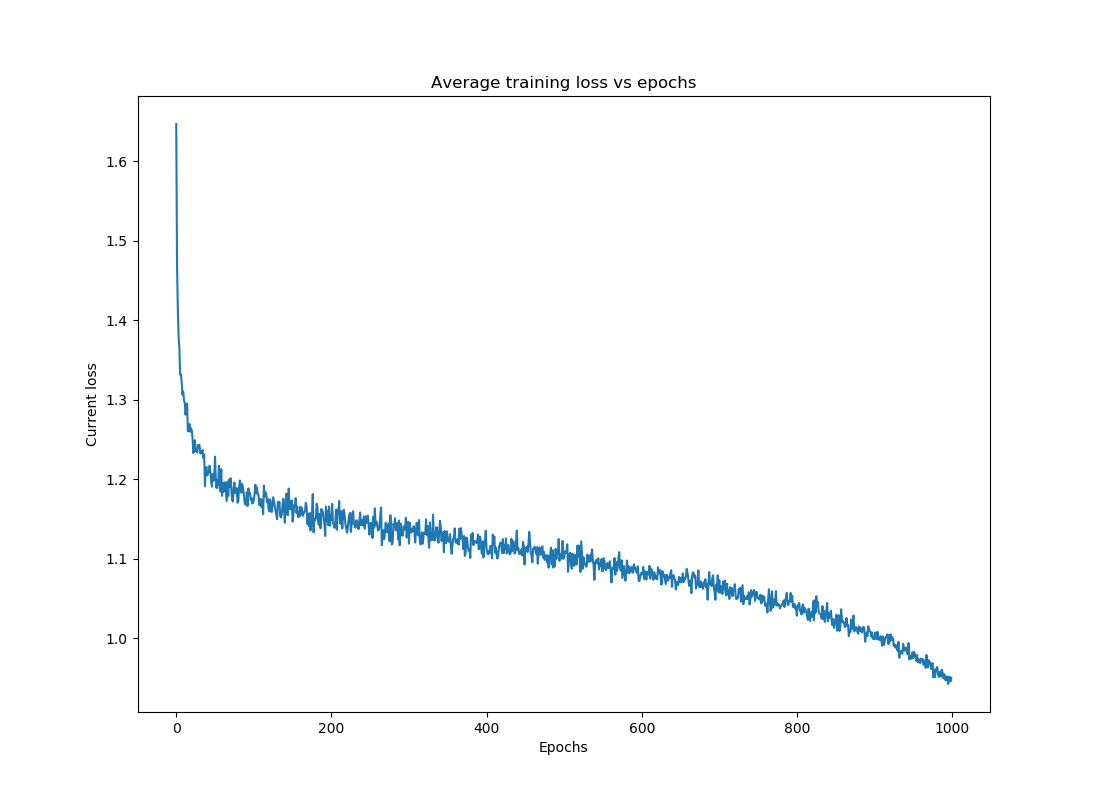}
        \includegraphics[width=10 cm, height=6.8 cm, trim=34 30 0 18]{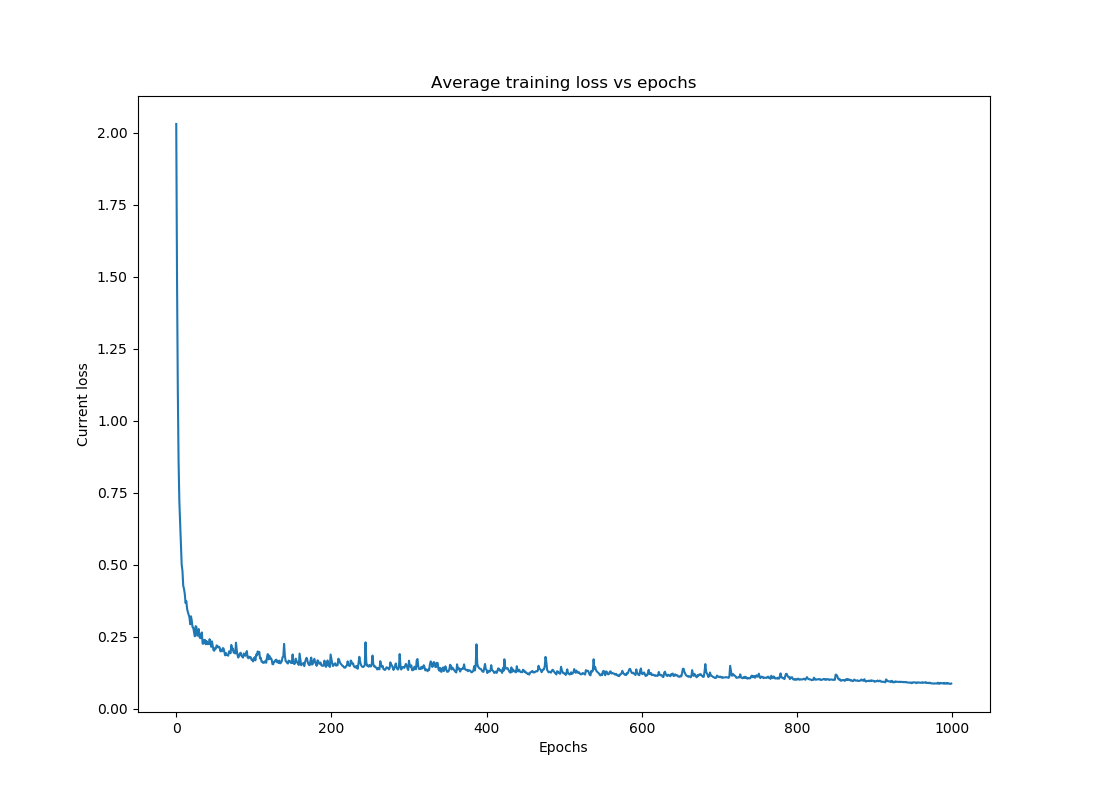}
    \caption{The convergence curve of our MSCFNet on Cityscapes (top) and CamVid (bottom) datasets. (Best viewed in color)}
\label{Figure 4}
\end{figure}

\begin{figure*}[t]
	\centering
	\includegraphics[width=18cm]{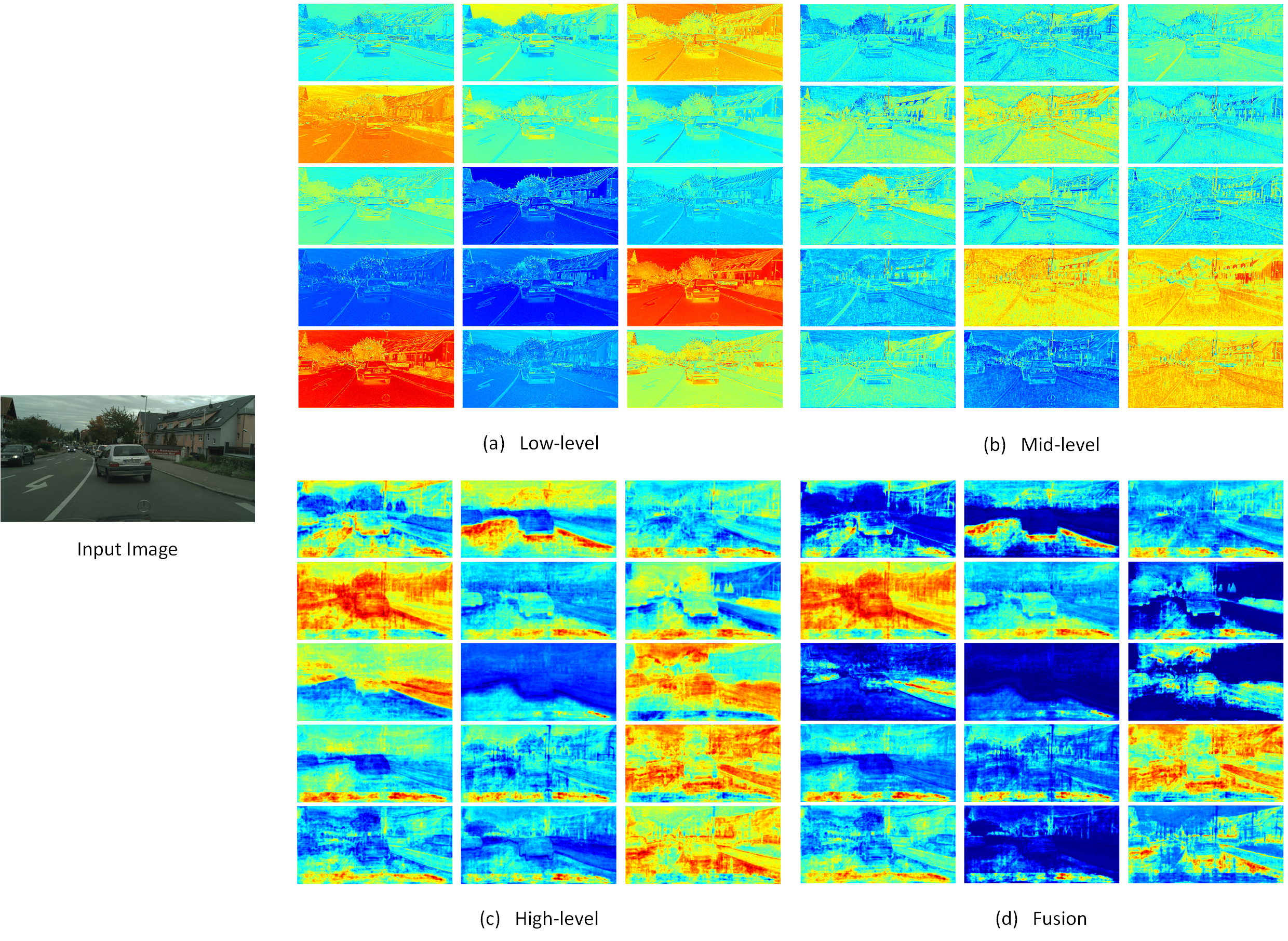}
	\caption{Feature maps with different levels on the Cityscapes validation set.}
	\label{Figure 5}
\end{figure*}

\subsection{Efficient Attention}
\label{sec32}

Generally, owing to the small number of network layers, a lightweight network can hardly extract deep enough features thoroughly like a large network. Consequently, producing representative features and combining them may be an essential manner to enhance the segmentation performance. To this end, we borrow the idea of the attention mechanism in our model. Attention is beneficial to both information integration and object feature emphasizing. Our efficient attention mechanisms include both spatial attention and channel attention modules.

\textbf{Spatial Attention.} The spatial attention we used is motivated by CBAM~\cite{woo2018cbam}, exploring the inter-spatial relationship of the input features to generate attention maps that depict where to highlight or suppress. The average pooling and max pooling operations are first adopted, and then concatenating them to formulate a feature descriptor followed by a convolution layer to produce the desired spatial weight maps. Finally, we multiply the attention maps and the input features of the module to obtain the final generated features. This procedure can be formulated as follows:

\begin{equation}
SA\left( F \right) =  {\sigma \left( {{f^{7 \times 7}}\left( {Concat\left[ {AvgP\left( F \right),MaxP\left( F \right)} \right]} \right)} \right) \times F},\
\label{eq5}
\end{equation}
where $F \in {R^{C \times H \times W}}$ denotes the input features, $\sigma$ is the sigmoid activation function, ${f^{7 \times 7}}$ denotes a standard convolution with the kernel size $7 \times 7$, $Concat$ means the concatenate operation, $AvgP$ and $MaxP$ represent the average pooling and max pooling operation, respectively.

The spatial attentions are placed between the three injection modules and the main branch (see Fig.~\ref{Figure 1}). The outputs of spatial attention are calculated in the following:

\begin{equation}
F_{sa}^n = SA\left( {\frac{1}{\beta }Input} \right),\beta  = {2^n},n = 1,2,3,
\label{eq6}
\end{equation}
where $Input$ is the observed image, $\beta$ denotes the factor of the reduction and $SA$ means the spatial attention operation.

\textbf{Channel Attention.} The channel attention we adapted is derived from ECANet~\cite{wang2020eca}, which just occupies a little computational resource but improves the performance greatly by comparison. We settle three channel attention modules followed by each feature combination operation as well as one before context fusion in the main branch, and two in the middle of the context branches. They can be simply separated into two categories, one is the attention operation in the encoder process, and the other is the attention operation in the decoder process (see Fig.~\ref{Figure 1}). The procedure of the above channel attention can be formulated as follows: 

\begin{equation}
CA\left( F \right) = \sigma \left( {{f^{k \times k}}\left( {T(AvgP(F))} \right)} \right) \times F,
\label{eq7}
\end{equation}
where $T$ represents the compression, transposition and extension operations of the tensor dimensions, ${f^{k \times k}}$ denotes a standard convolution with adaptive selection of kernel size $k$.

The attentive features of the encoding process can be calculated as follow:

\begin{equation}
F_{ca}^n = CA\left( {Concat\left( {{O_n},F_{sa}^n} \right)} \right),n = 1,2,3,
\label{eq8}
\end{equation}
where $F_{sa}^n$ and $F_{ca}^n$ represent the outputs of spatial attention and channel attention respectively, $O_1$ is the output of the initial block, $O_2$ and $O_3$ are the outputs of the two EAR blocks, $CA$ denotes the channel attention operation.

And the attentive features in the decoding process can be described as follow:

\begin{equation}
{F_{Bri}} = CA\left( {Up\left( {{I_i},\alpha } \right)} \right),\alpha  = {2^{i - 1}},i = 1,2,3,
\label{eq9}
\end{equation}
where $F_{Br1}$, $F_{Br2}$ and $F_{Br3}$ denote the output features of the main branch and the other two projection branches, $Up$ is the bilinear interpolate operation in the corresponding branches, $I_1$, $I_2$, $I_3$ indicate the low-level, intermediate-level, high-level contextual information respectively, and $\alpha$ is the magnification factor.

In summary, spatial attention focuses on indicating “where” to highlight while channel attention focuses on indicating “what” a given feature is. By considering these two attention mechanisms simultaneously at different stages in the network, our method can adaptively promote the representational power of the extracted features and facilitate the local and contextual information interaction greatly, which have been validated to be effective in the experimental section.

\begin{table*}[t]
	\caption{Ablation study results on CamVid testing set. CF: Context Fusion, FC: Fearure Concatenation, FA: Feature Adding, CA: Channel Attention, SA: Spatial Attention; R: Dilation Rate. Superscript'\dag' denotes the final version.}
	\begin{center}
		\setlength\tabcolsep{15pt}
		\begin{tabular}{@{}|l|ccc|c|c|@{}}
			\toprule[1pt]
			 &CF     &CA     &   &    &     \\ 

			Models							                &FC~~~~~~~~FA	               &SE~~~~~~~~ECA				 &SA             & mIoU (\%)         & Param 		\\	\hline\hline
			~(a) Context Fusion \\ \hline
			MSCFNet                                         &\XSolid~~~~~~~~\XSolid      &\XSolid~~~~~~~~\XSolid       &\XSolid        & 67.82            & 1.1429M \\ 
			MSCFNet 					                    &\Checkmark~~~~~~~~\XSolid   &\XSolid~~~~~~~~\XSolid       &\XSolid        & 68.42            & 1.1533M \\  
            MSCFNet 					                    &\XSolid~~~~~~~~\Checkmark   &\XSolid~~~~~~~~\XSolid       &\XSolid        & 68.87            & 1.1483M \\ \hline
			~(b) Attention Module\\ \hline
 			MSCFNet 					                    &\XSolid~~~~~~~~\Checkmark   &\Checkmark~~~~~~~~\XSolid    &\XSolid       & 68.96            & 1.2696M \\
			MSCFNet 					                    &\XSolid~~~~~~~~\Checkmark   &\XSolid~~~~~~~~\Checkmark    &\XSolid        & 69.16            & 1.1483M \\
			MSCFNet 					                    &\XSolid~~~~~~~~\Checkmark   &\XSolid~~~~~~~~\Checkmark    &\Checkmark     & 69.30            & 1.1486M \\ \hline
			~(c) Dilation Rate \\ \hline
			MSCFNet (R=1,1,2,1,2,5,7,9,17) 			        &\XSolid~~~~~~~~\Checkmark   &\XSolid~~~~~~~~\Checkmark    &\Checkmark     & 68.27            & 0.7684M \\
			MSCFNet (R=1,1,2,2,5,1,1,2,2,4,4,8,8,16,16)      &\XSolid~~~~~~~~\Checkmark   &\XSolid~~~~~~~~\Checkmark    &\Checkmark     & 69.04            & 1.1486M \\
			MSCFNet (R=1,1,2,2,5,1,2,5,7,9,2,5,7,9,17)\dag   &\XSolid~~~~~~~~\Checkmark   &\XSolid~~~~~~~~\Checkmark    &\Checkmark     & 69.30            & 1.1486M \\ \bottomrule[1pt]
		\end{tabular}
		\label{Table 1}
	\end{center}
\end{table*}

\begin{table}[t]
	\caption{Ablation study results by gradually adding intermediate contextual features on the CamVid testing set.}
	\begin{center}
		\setlength\tabcolsep{5pt}
		\begin{tabular}{@{}|l|c|c|@{}}
			\toprule[1pt]
			Models      & mIoU (\%)   &Param       \\ \hline\hline

			MSCFNet-High\underline{~}level                                                   & 67.91  &1.1432M           \\
			MSCFNet-High\underline{~}level + Mid\underline{~}level                           & 68.49  &1.1474M           \\
			MSCFNet-High\underline{~}level + Mid\underline{~}level + Low\underline{~}level   & 69.30  &1.1486M           \\ \bottomrule[1pt]

		\end{tabular}
		\label{Table 2}
	\end{center}
\end{table}

\begin{table}[t]
	\caption{Evaluation results on the Cityscapes testing set.}
	\begin{center}
		\setlength\tabcolsep{1pt}
		\begin{tabular}{@{}|l|c|c|c|c|c|@{}}
			\toprule[1pt]
			Methods                                & Pretrained & Input Size   & mIoU (\%)$\uparrow$     & Speed (FPS)$\uparrow$    & Param$\downarrow$    \\ \midrule\hline
			SegNet~\cite{badrinarayanan2017segnet} & ImageNet  & $640 \times 360$    & 56.1         & 17            & 29.50M \\
			ENet~\cite{paszke2016enet}             & No   & $512 \times 1024$         & 58.3         & 77            & \textbf{0.36M} \\
			FSSNet~\cite{8392426}     & -   & $512 \times 1024$          & 58.8         & 51            & - \\
			ESPNet~\cite{mehta2018espnet}          & No     & $512 \times 1024$       & 60.3         & 113           & \textbf{0.36M} \\
			CGNet~\cite{wu2020cgnet}               & No     & $1024 \times 2048$       & 64.8         & 50            & 0.50M \\
			NDNet~\cite{yang2020ndnet}			   & No   & $1024 \times 2048$         & 65.3         & 40            & 0.50M \\
			ContextNet~\cite{poudel2018contextnet} & No  & $1024 \times 2048$    & 66.1         & 18            & 0.85M \\
			EDANet~\cite{lo2019efficient}          & No    & $512 \times 1024$        & 67.3         & 81            & 0.68M \\
			ERFNet~\cite{romera2017erfnet}         & No   & $512 \times 1024$         & 68.0         & 42            & 2.10M \\
			Fast-SCNN~\cite{poudel2019fast}         & -     & $1024 \times 2048$        & 68.0         & 124  & 1.11M \\
			BiseNet~\cite{yu2018bisenet}           & ImageNet   & $768 \times 1536$   & 68.4         & 106 		  & 5.80M \\
			ICNet~\cite{zhao2018icnet}             & ImageNet  & $1024 \times 2048$    & 69.5         & 30 		      & 26.50M \\
			DABNet~\cite{li2019dabnet}             & No    & $1024 \times 2048$        & 70.1         & 28 		  	  & 0.80M \\
			FarSeeNet~\cite{zhang2020farsee}        & -     & $512 \times 1024$        & 70.2         & 69            & -     \\
			DFANet~\cite{li2019dfanet}          & ImageNet    & $512 \times 1024$  & 70.3         &\textbf{160} 		 	  & 7.80M     \\ 
			LEDNet~\cite{wang2019lednet}           & No   & $512 \times 1024$         & 70.6         & 71		 	  & 0.90M \\
            EdgeNet~\cite{han2020using}             & -    & $512 \times 1024$         & 71.0         & 31 		 	  & -     \\ \hline
			MSCFNet (ours)                         & No   & $512 \times 1024$         & \textbf{71.9}& 50 		      & 1.15M \\ \bottomrule[1pt]
		\end{tabular}
		\label{Table 3}
	\end{center}
\end{table}

\begin{table*}[t]
	\caption{Per-class IoU(\%) performance on the Cityscapes testing set. List of categories: Road, Sky, Car, Vegetation, Building, Side-walk, Pedestrian, Bus, Traffic Sign, Bicycle, Terrain, Traffic Light, Rider, Pole, Train, Motorcycle, Wall, Fence and Truck. 'Cla': 19 Classes.}
	\begin{center}
		\setlength\tabcolsep{4pt}
		\begin{tabular}{@{}|l|ccccccccccccccccccc|c|@{}}
			\toprule[1pt] 
			Methods                               & Roa  &Sky  &Car  &Veg  &Bui  &Sid  &Ped  &Bus  &TSi  &Bic  &Ter  &TLi  &Rid	 &Pol  &Tra  &Mot  &Wal	 &Fen  &Tru  &Cla   \\ \midrule\hline 
			SegNet~\cite{badrinarayanan2017segnet} & 96.4 &91.8 &89.3 &87.0 &84.0 &73.2 &62.8 &43.1 &45.1 &51.9 &63.8 &39.8 &42.8 &35.7 &44.1 &35.8 &28.4 &29.0 &38.1 &57.0  \\	        	
			ENet~\cite{paszke2016enet}             & 96.3 &90.6 &90.6 &88.6 &75.0 &74.2 &65.5 &50.5 &44.0 &55.4 &61.4 &34.1 &38.4 &43.4 &48.1 &38.8 &32.2 &33.2 &36.9 &58.3  \\
			ESPNet~\cite{mehta2018espnet}          & 97.0 &92.6 &92.3 &90.8 &76.2 &77.5 &67.0 &52.5 &46.3 &57.2 &63.2 &35.6 &40.9 &45.0 &50.1 &41.8 &35.0 &36.1 &38.1 &60.3  \\	        	
			CGNet~\cite{wu2020cgnet}               & 95.5 &92.9 &90.2 &89.6 &88.1 &78.7 &74.9 &59.5 &63.9 &60.2 &67.6 &59.8 &54.9 &54.1 &25.2 &47.3 &40.0 &43.0 &44.1 &64.8  \\	        	
			ERFNet~\cite{romera2017erfnet}         & 97.7 &94.2 &92.8 &91.4 &89.8 &81.0 &76.8 &60.1 &65.3 &61.7 &68.2 &59.8 &57.1 &56.3 &51.8 &47.3 &42.5 &48.0 &50.8 &68.0  \\	        	
			ICNet~\cite{zhao2018icnet}             & 97.1 &93.5 &92.6 &91.5 &89.7 &79.2 &74.6 &\textbf{72.7}	&63.4 &70.5 &68.3 &60.4	&56.1 &61.5 &51.3 &53.6	&43.2 &48.9	&51.3 &69.5  \\
			DABNet~\cite{li2019dabnet}             & 97.9 &92.8 &93.7 &91.8 &90.6 &82.0 &78.1 &63.7 &67.7 &66.8 &70.1 &63.5 &57.8 &59.3 &56.0 &51.3 &45.5 &50.1 &52.8 &70.1  \\
			LEDNet~\cite{wang2019lednet}           & \textbf{98.1} &\textbf{94.9} &90.9 &\textbf{92.6} &\textbf{91.6} &79.5 &76.2 &64.0 &\textbf{72.8} &\textbf{71.6} &61.2 &61.3 &53.7 &\textbf{62.8} &\textbf{52.7} &44.4 &47.7 &49.9 &\textbf{64.4} &70.6 \\
            EdgeNet~\cite{han2020using}            & \textbf{98.1} &\textbf{94.9} &\textbf{94.3} &92.4 &\textbf{91.6} &\textbf{83.1} &80.4 &60.9 &71.4 &67.7 &69.7 &\textbf{67.2} &61.1 &62.6 &52.5 &55.3 &45.4 &50.6 &50.0  &71.0  \\\hline
			MSCFNet (ours)                         & 97.7 &94.3 &94.1 &92.3 &91.0 &82.8 &\textbf{82.7} &66.1 &71.4 &70.2 &\textbf{70.2} &67.1 &\textbf{62.7} &61.2 &51.9 &\textbf{57.6} &\textbf{49.0} &\textbf{52.5} &50.9 &\textbf{71.9}  \\\bottomrule[1pt]
		\end{tabular}
		\label{Table 4}
	\end{center}
\end{table*}

\begin{table}[t]
	\caption{Comparison results on the CamVid testing set.}
	\begin{center}
		\setlength\tabcolsep{10pt}
		\begin{tabular}{@{}|l|c|c|c|@{}}
			\toprule[1pt]
			Methods  & Pretrained   & mIoU (\%)$\uparrow$        & Param$\downarrow$    \\ \midrule\hline
			ENet~\cite{paszke2016enet}              & No  & 51.3            & \textbf{0.36M} \\
			SegNet~\cite{badrinarayanan2017segnet}   & ImageNet & 55.6            & 29.50M \\
			FCN-8s~\cite{long2015fully}              & ImageNet & 57.0            & 134.50M \\
			NDNet~\cite{yang2020ndnet}               & No  & 57.2            & 0.50M   \\
			DeepLabLFOV~\cite{chen2014semantic}      & ImageNet & 61.6            & 37.30M \\
			DFANet~\cite{li2019dfanet}               & ImageNet  & 64.7            & 7.80M \\ 
			Dilation8~\cite{yu2015multi}             & ImageNet & 65.3            & 140.80M \\
			CGNet~\cite{wu2020cgnet}                 & No & 65.6            & 0.50M \\
			BiseNet~\cite{yu2018bisenet}            & ImageNet  & 65.6        	& 5.80M \\
		 	DABNet~\cite{li2019dabnet}              & No  & 66.4            & 0.76M \\
			ICNet~\cite{zhao2018icnet}              & ImageNet  & 67.1            & 26.50M \\ \hline
			MSCFNet (ours)                           & No & \textbf{69.3}   & 1.15M \\ \bottomrule[1pt]
		\end{tabular}
		\label{Table 5}
	\end{center}
\end{table}

\subsection{Context Fusion}
\label{sec33}

Previous methods usually learnt finer-scale predictions in a stage-by-stage manner. That means, the net in each stage is trained by the initialization of the previous stage net, which may result in the contextual cues cannot complement each other. Meanwhile, some remarkable networks have shown that good performance usually stems from a fusion of hierarchical information. Hence, we adopt this idea in our method before the final classification to integrate the multi-scale contextual information. The features $F_{cf}$ from the context fusion  operation can be formulated as follow:

\begin{equation}
{F_{cf}} = \rho \left( {{F_{Br1}} + {F_{Br2}} + {F_{Br3}}} \right).
\label{eq10}
\end{equation}

Therefore, it can be seen that the context fusion module combining attentive contextual information from different stages of the network, which can alleviate the limitations caused by the spatial statistics of pixels loss.

\subsection{Network Architecture}
\label{sec34}

In this subsection, we introduce the entire lightweight network as presented in Fig.~\ref{Figure 1}. MSCFNet has an asymmetric structure with an encoder structure and the related decoder structure, finally followed by a widely used classification layer.

\textbf{Encoder.} In the main flow, the initial feature extraction module includes three $3 \times 3$ convolutions, in which the first one uses stride 2 to extract feature information and reduce the size simultaneously. Then the downsampling block we employed has two alternative outputs of a $3 \times 3$ convolution with stride 2 and a $2 \times 2$ max-pooling with stride 2. If the amount of the input channels is larger than or equal to that of the output channels, the block is just the single $3 \times 3$ convolution. Otherwise, the max-pooling operation is added, the concatenation of these two branches forms the final downsampling outputs. Please see Fig.~\ref{Figure 3} for more details~\cite{lo2019efficient}.

The downsampling operation amplifies the receptive field for collecting more contextual information. Nevertheless, the reduction in terms of the resolution of the input feature maps often leads to spatial and boundary resolution loss. By taking these into account, we just perform downsampling three times progressively and obtain 1/8 resolution of the original feature map to gather deeper context but maintain more image details. Followed each downsampling block is the EAR block, which includes different numbers of consecutive EAR modules. The first block has 5 EAR modules, while the second consists of 10 EAR modules for dense feature extraction. To better promote feature propagation and contextual information relationship, we apply inter-block concatenation to combine high-level and low-level features. Also, we employ dilation convolution in the blocks as depicted in Section~\ref{sec22}. The dilation rates in the first block are $\left\{ {1,1,2,2,5} \right\}$, and the second are $\left\{ {1,2,5,7,9,2,5,7,9,17} \right\}$, respectively. We choose this scheme to mitigate the gridding artifacts and enlarge the receptive field for more context of larger scope. 

For better feature reuse, we insert efficient spatial attention in the shortcut connections between the input features, which is handled by three input injection modules with $\rm{1/2}$, $\rm{1/4}$, $\rm{1/8}$ ratios, and two downsampling blocks, as well as the last convolution layer respectively. Moreover, to integrate contextual semantic information and allocate channel information resources of the feature maps, we use effective channel attention modules followed the above three identical places of the feature concatenations. What’s more, we apply long-range shortcut connections in our network. The two context branches are composed of efficient channel attentions covering contextual information from different scales. Also, this operation alleviates the contradiction that shallow features lack semantic information and deep features lack boundary and detailed information. 

\textbf{Decoder.} Our Decoder is asymmetrical relative to the Encoder. It entirely uses a $\times 4$ upsampling and a final $\times 2$ deconvolution operations to restore the original input image size and output the final segmentation prediction simultaneously. Compared to some existing networks directly upsampling 8 times or others decoding step by step, our two-step decoder structure combines the superiority of the two, which not only ensures the simplicity of the calculation, but also guarantees the maximum recovery of the decoded information.

\section{Experiments}
\label{secaa4}

\subsection{Implementation Protocol}
\label{sec41}

\textbf{Datasets.} We use two popular benchmarks of urban street scenes -- Cityscapes~\cite{cordts2016cityscapes} and CamVid to assess the effectiveness of our proposed network. The Cityscapes dataset contains 19 semantic categories including 5000 fine-annotated samples with the size of $2048 \times 1024$ that are separated into three subsets: 2975 samples used for training, 500 samples used for validation, and the other 1525 samples used for testing. The CamVid is another smaller dataset for self-driving scenarios. It has 11 semantic categories with 367 training, 101 validation, and 233 testing samples, of which the image size is $960 \times 720$. For Cityscapes and CamVid, the original input is resized to $1024 \times 512$ and $480 \times 360$ for network training, respectively.

\textbf{Parameter Settings.} For the Cityscapes dataset, we train our network end-to-end by exploiting the stochastic gradient descent (SGD)~\cite{bottou2010large} method with a batch size of 4 to leverage the hardware memory. The momentum is set as 0.9 and also the related weight decay is set as $1 \times 10^{-4}$. The ``poly'' policy for learning rate is also adopted, where the learning rate is adaptively adjusted according to the following equation after every iteration:

\begin{equation}
lr=lr_{base}\times\left(1-\frac{iteration}{max\_iteration}\right)^{power},\\
\label{eq11}
\end{equation}
where $lr_{base}$ is the initial learning rate, $lr$ is the learning rate after each iteration, $iteration$ is the index of the current iteration, $max\_iteration$ is the maximum number of iterations in each epoch. We configure the initial learning rate as $4.5 \times 10^{-2}$ and the power is 0.9.

When performing training on the CamVid dataset, we adjust the optimization method to Adam with a batch size of 8. The momentum is set as 0.9 and the weight decay is set as $2 \times 10^{-4}$. Also, we employ the ``poly'' learning rate policy with the initial learning rate of $1 \times 10^{-3}$. Fig.~\ref{Figure 4} plots the curves of the loss function vs the number of iterations on the Cityscapes and CamVid datasets. The two curves drop smoothly and converge eventually, indicating that our MSCFNet can be well trained.

\begin{figure*}[t]
	\centerline{\includegraphics[width=18cm]{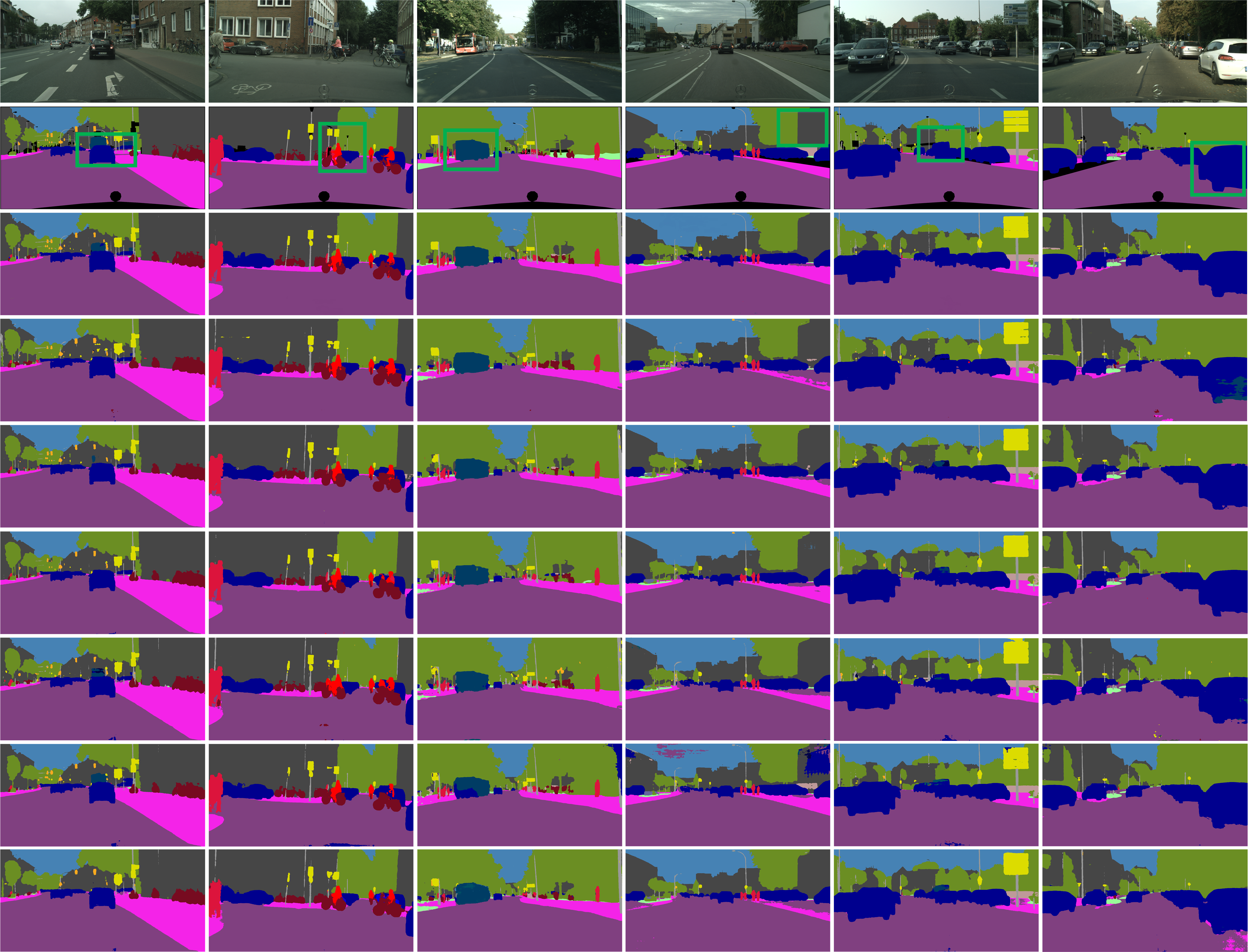}}
	\caption{The comparative results on the Cityscapes val dataset. From top to the bottom are successively the original observed images, ground truths, segmentation results from our MSCFNet,  LEDNet~\cite{wang2019lednet}, DABNet~\cite{li2019dabnet}, ERFNet~\cite{romera2017erfnet}, EDANet~\cite{lo2019efficient}, CGNet~\cite{wu2020cgnet} and ESPNet~\cite{mehta2018espnet}. (Best viewed in color)}
	\label{Figure 6}
\end{figure*}

\subsection{Ablation Study}
\label{sec42}

In this part, we conduct comparative studies to demonstrate the feasibility and effectiveness of our presented method. All the ablation studies are conducted on the CamVid training set, validation set, and evaluated on its testing set.

\textbf{Context Fusion.} To study how the contextual features affect the segmentation accuracy, we have performed experiments without any attention mechanism in this part. From Table~\ref{Table 1} (a) we could observe that the segmentation results without the context fusion mechanism are more than 1\% lower than those used, which has confirmed that multi-scale contextual features play a critical role in dense pixel classification tasks. As for the way of context fusion, feature adding has better performance than that of feature concatenation. In Table~\ref{Table 2}, we also study how the different levels of features affect the segmentation results. Fig.~\ref{Figure 5} depicts the heat maps from the low-level, mid-level, and high-level contextual features.

\textbf{Attention Module.} We use the attention mechanism for better extracting spatial features and promoting channel information interaction. Table~\ref{Table 1} (b) demonstrates the effectiveness brought by channel and spatial attention. Adding channel attention SE~\cite{hu2018squeeze} can bring 0.1\% slightly better accuracy. However, ECA~\cite{wang2020eca} works better with fewer parameters. When we add spatial attention to the injection branches, the segmentation accuracy increases from 69.16\% to 69.30\%. From the above studies, we conclude that by using the channel and spatial attention mechanisms simultaneously, we can gain better segmentation performance at the cost of negligible parameters increasing.

\textbf{Dilation Rate.} As depicted in Table~\ref{Table 1} (c), we design three experiments to study the effect of the number of EAR modules and the dilation rate. First, we reduce the EAR modules to 3 in the first block and to 6 in the second block. We find that the segmentation accuracy is 1.03\% lower than that of our final version, indicating that more EAR modules can improve the performance. Then setting the same number of modules, we adopt the idea of DABNet~\cite{li2019dabnet} with the dilation rates are the power of 2 in the second EAR block. The results also validate that our settings can achieve better performance.

\begin{figure*}[t]
	\centerline{\includegraphics[width=18cm]{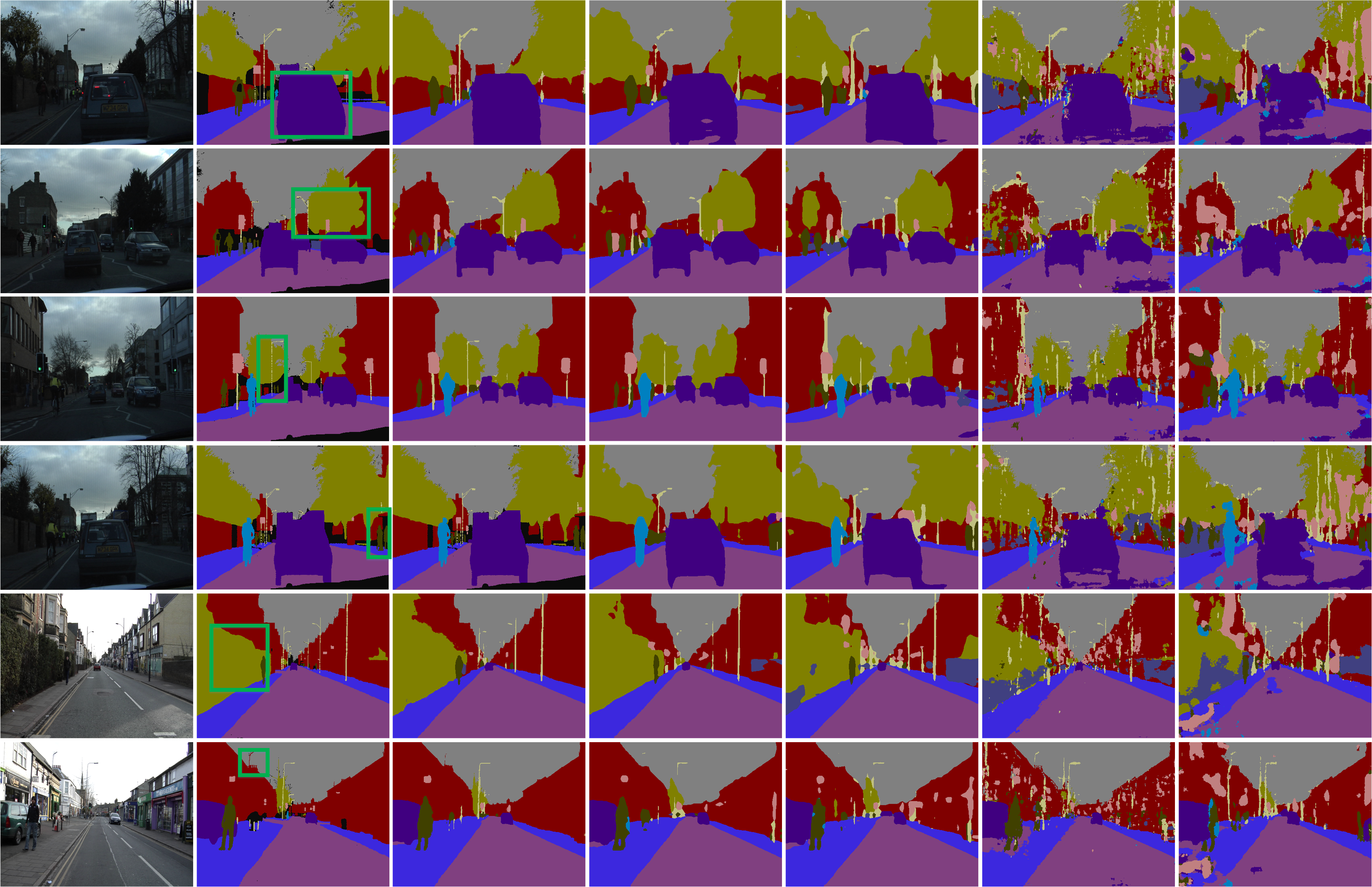}}
	\caption{The comparative results on the CamVid testing set. From left to right are original observed images, ground truths, segmentation outputs from our MSCFNet,  DABNet~\cite{li2019dabnet}, CGNet~\cite{wu2020cgnet}, SegNet~\cite{badrinarayanan2017segnet} and ENet~\cite{paszke2016enet}. (Best viewed in color)}
	\label{Figure 7}
\end{figure*}

\subsection{Evaluation Results}

The performance of our MSCFNet is evaluated with several state-of-the-art ones in this part on the above mentioned Cityscapes and CamVid datasets: FCN-8s~\cite{long2015fully}, DeepLabLFOV~\cite{chen2014semantic}, Dilation8~\cite{yu2015multi}, ENet~\cite{paszke2016enet}, FSSNet~\cite{8392426}, SegNet~\cite{badrinarayanan2017segnet}, ERFNet~\cite{romera2017erfnet}, Fast-SCNN~\cite{poudel2019fast}, ESPNet~\cite{mehta2018espnet}, CGNet~\cite{wu2020cgnet}, NDNet~\cite{yang2020ndnet}, ContextNet~\cite{poudel2018contextnet}, ICNet~\cite{zhao2018icnet}, BiseNet~\cite{yu2018bisenet}, EDANet~\cite{lo2019efficient}, DABNet~\cite{li2019dabnet}, FarSeeNet~\cite{zhang2020farsee}, LEDNet~\cite{wang2019lednet}, DFANet~\cite{li2019dfanet}, and EdgeNet~\cite{han2020using}. 

As can be observed from Table~\ref{Table 3} to Table~\ref{Table 5}, the comparison results verify that our MSCFNet attains a better balance between segmentation accuracy and efficiency. For the Cityscapes dataset, our MSCFNet only has a 1.15M model size but yields 71.9\% class mIoU and 88.4\% category mIoU, respectively, even 74.2\% class mIoU on the validation set. Considering the efficiency, MSCFNet only occupies 15\% of the number of parameters of the ICNet but achieves a faster speed and a better result. Although DABNet is almost 0.4M smaller than our network, it delivers poor accuracy with 1.8\% lower than our MSCFNet. For the CamVid dataset, we can see that our model also gets remarkable performance with a smaller capacity, and achieves 69.3\% class mIoU on the CamVid testing set, which is superior to most of the existing competitive methods. Although DFANet has a faster speed, it occupies almost 7$\times$ number of parameters than our MSCFNet. As shown in these tables, our network makes a good trade-off among model size, inference speed, and segmentation accuracy. The qualitative comparisons with some respective methods are also shown in Fig.~\ref{Figure 6} and Fig.~\ref{Figure 7}, which qualitatively verify the effectiveness of our MSCFNet.

\section{Conclusions}

In summary, we have proposed a multi-scale context fusion network (MSCFNet), which improves both segmentation accuracy and inference speed for lightweight semantic segmentation task in this work. We designed an efficient asymmetric residual (EAR) module, which adopts factorization depth-wise convolution and dilation convolution to capture object features with a lower number of parameters and computational budgets using different receptive fields. Moreover, MSCFNet had branches with efficient attention modules from different stages extracting multi-scale contextual information. Then, the features from these connections were combined to enhance the expression of the features and facilitate the local and contextual information interaction greatly. Extensive experiments on the CamVid and Cityscapes datasets have validated that our architecture can attain a better balance between efficiency and accuracy than several comparative approaches.

\bibliographystyle{IEEEtran}
\bibliography{reference}

%
\begin{IEEEbiography}[{\includegraphics[width=1in,height=1.25in,clip,keepaspectratio]{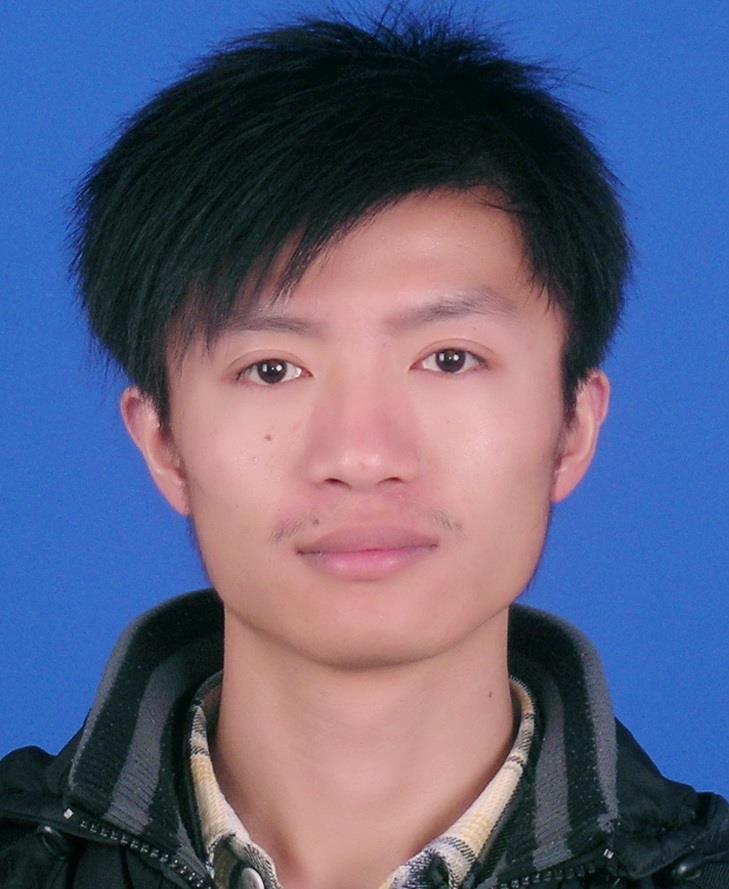}}]{Guangwei Gao}
(M’17) received the Ph.D. degree in pattern recognition and intelligence systems from the Nanjing University of Science and Technology, Nanjing, in 2014. He was an Exchange Student of the Department of Computing, The Hong Kong Polytechnic University, in 2011 and 2013, respectively. He was a Project Researcher with the National Institute of Informatics, Tokyo, Japan, in 2019. He is currently an Associate Professor with the Institute of Advanced Technology, Nanjing University of Posts and Telecommunications. His research interests include pattern recognition, and computer vision. He has served as reviewer for IEEE TNNLS/TIP/TMM/TCYB, Pattern Recognition, Neurocomputing, Pattern Recognition Letters, and AAAI/ICPR/ICIP etc.
\end{IEEEbiography}

\begin{IEEEbiography}[{\includegraphics[width=1in,height=1.25in,clip,keepaspectratio]{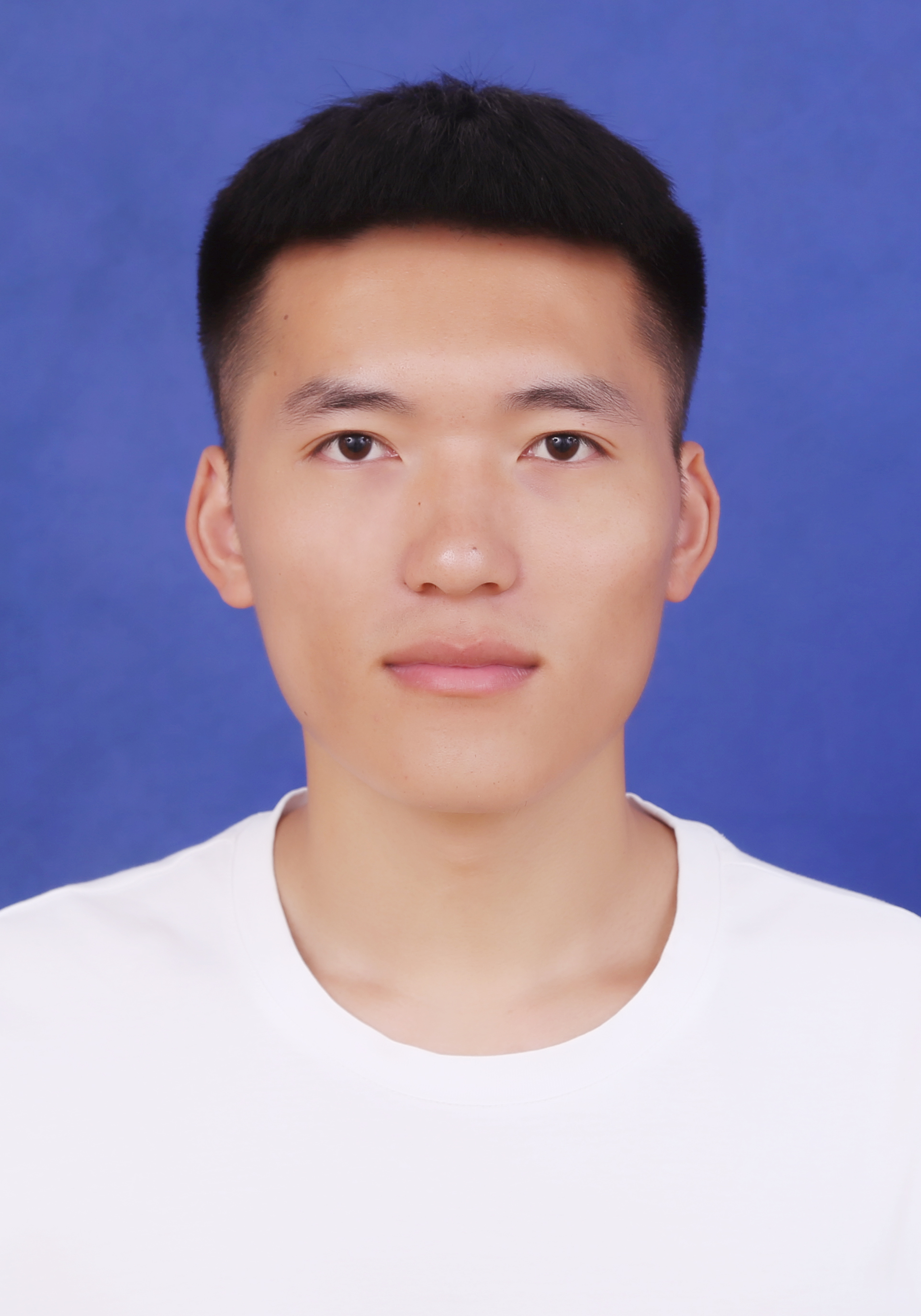}}]{Guoan xu}
received the B.S degrees in Measurement Control Technology and Instrumentation from Changshu Institute of Technology, Jiangsu, China, in 2019. He is currently pursuing the M.S. degree with the College of Automation \& College of Artificial Intelligence, Nanjing University of Posts and Telecommunications. His research interests include image semantic segmentation.
\end{IEEEbiography}

\begin{IEEEbiography}[{\includegraphics[width=1in,height=1.25in,clip,keepaspectratio]{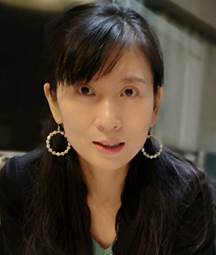}}]{Yi Yu}
(Member, IEEE) received the Ph.D. degree in information and computer science from Nara Women’s University, Japan. He is currently an Assistant Professor with the National Institute of Informatics (NII), Japan. Before joining NII, she was a Senior Research Fellow with the School of Computing, National University of Singapore. Her research covers large-scale multimedia data mining and pattern analysis, location-based mobile media service and social media analysis. She and her team received the best Paper Award from the IEEE ISM 2012, the 2nd prize in Yahoo Flickr Grand Challenge 2015, were in the top winners (out of 29 teams) from ACM SIGSPATIAL GIS Cup 2013, and the Best Paper Runner-Up in APWeb-WAIM 2017, recognized as finalist of the World’s FIRST 10K Best Paper Award in ICME 2017.
\end{IEEEbiography}

\begin{IEEEbiography}[{\includegraphics[width=1in,height=1.25in,clip,keepaspectratio]{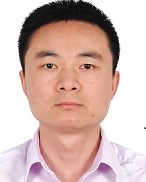}}]{Jin Xie}
(Member, IEEE) received the Ph.D degree from the Department of Computing, Hong Kong Polytechnic University in 2012. From 2013 to 2017, he was a Research Scientist with New York University Abu Dhabi, Abu Dhabi, UAE. Now, he is a professor in the School of Computer Science and Engineering of Nanjing University of Science and Technology. His current research interests include image forensics, computer vision, machine learning, 3-D computer vision with the convex optimization, and deep learning methods.
\end{IEEEbiography}

\begin{IEEEbiography}[{\includegraphics[width=1in,height=1.25in,clip,keepaspectratio]{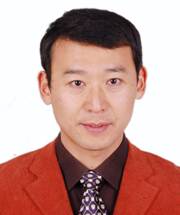}}]{Jian Yang}
(Member, IEEE) received the PhD degree from Nanjing University of Science and Technology (NUST), on the subject of pattern recognition and intelligence systems in 2002. In 2003, he was a postdoctoral researcher at the University of Zaragoza. From 2004 to 2006, he was a Postdoctoral Fellow at Biometrics Centre of Hong Kong Polytechnic University. From 2006 to 2007, he was a Postdoctoral Fellow at Department of Computer Science of New Jersey Institute of Technology. Now, he is a Chang-Jiang professor in the School of Computer Science and Engineering of NUST. He is the author of more than 100 scientific papers in pattern recognition and computer vision. His papers have been cited more than 4000 times in the Web of Science, and 9000 times in the Scholar Google. His research interests include pattern recognition, computer vision and machine learning. Currently, he is/was an Associate Editor of Pattern Recognition Letters, IEEE Trans. Neural Networks and Learning Systems, and Neurocomputing. He is a Fellow of IAPR.
\end{IEEEbiography}

\begin{IEEEbiography}[{\includegraphics[width=1in,height=1.25in,clip,keepaspectratio]{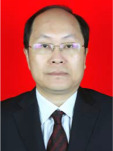}}]{Dong Yue}
(Fellow, IEEE) received the Ph.D. degree in engineering from the South China University of Technology, Guangzhou, China, in 1995. He is currently a Professor and Dean of the Institute of Advanced Technology and College of Automation \& AI, Nanjing University of Posts and Telecommunications. He was the Associate Editor for the Journal of the Franklin Institute and International Journal of Systems Sciences and the Guest Editor of Special Issue on New Trends in Energy Internet: Artificial Intelligence-based Control, Network Security and Management. Up to now, he has authored or co-authored more than 250 papers in international journals and two books. He holds more than 50 patents. His current research interests include analysis and synthesis of networked control systems, multiagent systems, optimal control of power systems, and Internet of Things. Prof. Yue served as the Associate Editor for IEEE Industrial Electronics Magazine, IEEE Transactions on Industrial Informatics, IEEE Transactions on Systems, Man and Cybernetics: Systems, IEEE Transactions on Neural Networks and Learning Systems.
\end{IEEEbiography}




\end{document}